\documentclass[10pt,twocolumn,letterpaper]{article}

\usepackage{wacv}              % To produce the CAMERA-READY version
\usepackage{graphicx}
\usepackage{amsmath}
\usepackage{amssymb}
\usepackage{booktabs}
\usepackage{bm}
\usepackage{multirow}
\usepackage{soul}

\usepackage[pagebackref,breaklinks,colorlinks]{hyperref}

\usepackage[capitalize]{cleveref}
\crefname{section}{Sec.}{Secs.}
\Crefname{section}{Section}{Sections}
\Crefname{table}{Table}{Tables}
\crefname{table}{Tab.}{Tabs.}

\def\wacvPaperID{10} % *** Enter the WACV Paper ID here
\def\confName{WACV}
\def\confYear{2025}

\begin{document}

%%%%%%%%% TITLE - PLEASE UPDATE
\title{DiffFake: Exposing Deepfakes using Differential Anomaly Detection}

\author{Sotirios Stamnas, Victor Sanchez\\ 
University of Warwick\\
{\tt\small \{Sotirios.Stamnas, v.f.sanchez-silva\}@warwick.ac.uk}
% For a paper whose authors are all at the same institution,
% omit the following lines up until the closing ``}''.
% Additional authors and addresses can be added with ``\and'',
% just like the second author.
% To save space, use either the email address or home page, not both
}
\maketitle

%%%%%%%%% ABSTRACT
\begin{abstract}
   Traditional deepfake detectors have dealt with the detection problem as a binary classification task. This approach can achieve satisfactory results in cases where samples of a given deepfake generation technique have been seen during training, but can easily fail with deepfakes generated by other techniques. In this paper, we propose DiffFake, a novel deepfake detector that approaches the detection problem as an anomaly detection task. Specifically, DiffFake learns natural changes that occur between two facial images of the same person by leveraging a differential anomaly detection framework. This is done by combining pairs of deep face embeddings and using them to train an anomaly detection model. We further propose to train a feature extractor on pseudo-deepfakes with global and local artifacts, to extract meaningful and generalizable features that can then be used to train the anomaly detection model. We perform extensive experiments on five different deepfake datasets and show that our method can match and sometimes even exceed the performance of state-of-the-art competitors.
\end{abstract}

%%%%%%%%% BODY TEXT
\section{Introduction}
\label{sec:intro}

The term \textit{deepfake} refers to videos or images that have been manipulated to depict real or non-existent people, often with malicious intent. These media pose an increasing threat as deep learning techniques, such as generative adversarial networks (GANs) \cite{goodfellow2014GAN,karras2019styleGAN} and diffusion models \cite{ho2020DDPM,rombach2022StableDiff} have rapidly developed, enabling the creation of deepfakes indistinguishable from authentic media. Two of the most common deepfake manipulations in videos include face swap (FS), and facial reenactment (FR), where a person's identity or facial expressions can be altered. Deepfakes can have serious implications for security, as they can be used to spread misinformation and infringe on privacy \cite{dickson2021deepfake,oltermann2022european}. 

Given this threat, the research community has developed many deep-learning-based methods to detect deepfakes. Early work on deepfake detection has formulated the problem as a binary classification task, where a deep neural network is trained on both real and fake media in a supervised fashion. These methods achieve excellent performance in in-dataset scenarios where the manipulation methods encountered during testing are also present in the training. However, their performance can drastically decrease in two critical settings: (1) \textit{cross-manipulation} scenarios where a model has been trained on a specific manipulation type and tested on another, \eg trained on FS and tested on FR; and (2) \textit{cross-dataset} scenarios where the manipulation methods  that are encountered during testing (potentially generating the same type of manipulation) are not seen during training. These generalization issues are the primary challenges in deepfake detection research, as in real-world applications the source or type of a given manipulation is often unknown. Therefore, constructing a highly robust and generalizable deepfake detector is highly important in the deepfake detection community.

To address this issue, recent work on deepfake detection has focused on developing methods that can learn more generalizable features \cite{zhang2019detecting_freq,haliassos2021lips,maiano2022depthfake,amerini2019optical_flow}. One of the most effective approaches is to use dedicated data-augmentation techniques to generate synthetic images that simulate common artifacts present in deepfakes, such as blending boundaries \cite{li2020face_x_ray}, inconsistencies in the frequency domain \cite{chen2021local_freq}, and color mismatch \cite{shiohara2022detecting}. These artifacts can be present either in the entire face (global) or in specific regions of the face (local). These synthetic images, referred to in the literature as \textit{pseudo-deepfakes}, can then be used to train a classifier with a much greater generalization capability than traditional methods. 

\begin{figure}[t]
    \centering
    \includegraphics[width=1.0\linewidth]{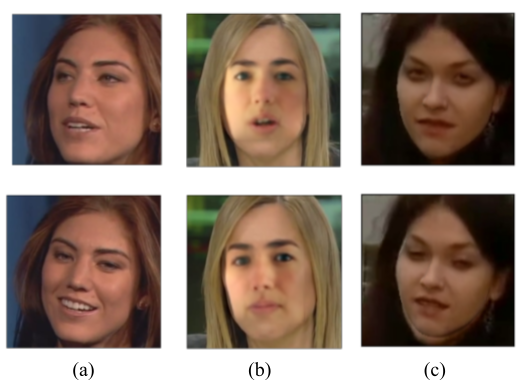}
    \caption{Example of changes that occur between frames of real and fake videos. (a) corresponds to frames from a real video that exhibit a natural change between the two head poses. (b) corresponds to a deepfake video and exhibits illumination inconsistency on the left side of the face. (c) corresponds to a deepfake video and exhibits facial boundary inconsistency around the chin and jawline regions.}
    \label{fig:unnatural_changes}
\end{figure}

This paper proposes DiffFake a novel approach for detecting deepfakes that combines pseudo-deepfake generation with anomaly detection. Specifically, we introduce a differential anomaly detection framework \cite{scherhag2020deep_diff_AD,diff_AD}, which allows learning natural changes between two real facial images of the same person. The motivation behind this idea is that deepfake videos usually exhibit unnatural changes in the facial region of a given person, as shown in Fig.~\ref{fig:unnatural_changes}. Therefore, the goal is to detect such cases as anomalies. Firstly, our method involves generating pseudo-deepfakes, which are then used to train a backbone to extract meaningful features from facial images. Unlike other competing methods that generate pseudo-deepfakes with either local or global artifacts, we propose a mask generation scheme that introduces both local and global artifacts, further enriching the discriminative capabilities of the backbone. The backbone is then used to extract feature vectors from pairs of images featuring the same person, which are combined and used to train an anomaly detection model (ADM). Unlike other anomaly detection-based methods for deepfake detection, which rely on information extracted from a single frame \eg \cite{khalid2020ocfakedect,larue2023seeable,mejri2023untag,leyva2024data}, our method captures information from pairs of images, which enhances its generalization capabilities, leading to competitive results. % as we will show later.

We demonstrate the competitive performance of DiffFake through extensive experiments on five different open-source deepfake detection datasets, \ie, FF++ \cite{rossler2019faceforensics++}, CDF \cite{li2020celeb}, DF1.0 \cite{jiang2020deeperforensics}, FNet \cite{he2021forgerynet} and FSh \cite{li2020FaceShifter}. Our experiments include three of the most relevant experimental settings in deepfake detection: (1) cross-manipulation scenario, (2) cross-dataset scenario, and (3) degrading video quality scenario. The experimental results show that our method achieves competitive performance across all datasets and in some cases even outperforms state-of-the-art (SoTA) competitors.

The main contributions of our paper are: 
\begin{itemize}
    \item We propose DiffFake, a novel deepfake detector based on a differential anomaly detection framework. Unlike existing AD-based deepfake detection techniques that use information from individual frames, DiffFake learns natural changes that occur in pairs of images of the same subject.
    \item We introduce a data-augmentation technique that generates pseudo-deepfakes with both local and global artifacts, using a facial landmark-based mask generation scheme. The pseudo-deepfakes are then used along with real images to train a backbone, allowing for the extraction of generalizable feature vectors for our AD model.
    \item We perform rigorous experiments under three experimental settings (cross-manipulation, cross-dataset, and degrading video quality) which demonstrate the competitive performance of our approach.
\end{itemize}
\section{Related work}

Owing to the wide variety of approaches to the deepfake detection task, in this section, we provide an overview of the most relevant methods, laying the groundwork for introducing DiffFake.

\noindent\textbf{Binary classification approaches.} Most early work on deepfake detection deals with the problem as a binary classification task, where both real and fake media are used to train a classifier model. These methods use a variety of different network architectures such as constrained layers \cite{bayar2016deep_constrained_layers}, shallow networks \cite{afchar2018mesonet}, depthwise convolution layers \cite{rossler2019faceforensics++}, networks with attention mechanisms \cite{dang2020detection_attention,zhao2021multi}, and recurrent convolutional networks \cite{sabir2019recurrent,guera2018deepfake_recurrent}, which try to leverage temporal inconsistencies between subsequent video frames. These methods perform considerably well in the in-dataset setting but mostly fail in dealing with unseen deepfake generators. To address this generalization issue, subsequent work on deepfake detection has focused on leveraging specific representations to capture forgery traces more effectively. Such representations extract information from eye blinking \cite{jung2020blinking}, head poses \cite{yang2019head_poses}, mouth movement \cite{haliassos2021lips}, optical flow \cite{amerini2019optical_flow}, and depth-maps \cite{maiano2022depthfake}. Furthermore, several methods have focused on extracting information from the frequency domain \cite{zhang2019detecting_freq,frank2020leveraging_freq,qian2020thinking_in_freq,liu2021spatial_phase_freq,luo2021generalizing_freq}, which has also been shown to improve generalization capabilities. However, all of these methods rely on using both real and fake media for training, which can result in overfitting in specific manipulation types or methods that are present in the training set.

\noindent\textbf{Pseudo-deepfake generation.} Arguably one of the most popular and effective methods for generalizable deepfake detection is to make use of dedicated data augmentation techniques, that leverage only real images, to synthesize so-called \textit{pseudo-deepfakes}, which contain common artifacts found in actual deepfakes. In the case of images or frames depicting faces, this process broadly involves blending a person's face from a source image to another person's face in a given target image. This idea is first introduced in Face X-ray \cite{li2020face_x_ray}, where blended faces are generated by using images of different subjects for the source and target. A main drawback of Face X-ray is the use of a nearest landmark search for source-target pair selection, which can be computationally expensive. Shiohara \etal \cite{shiohara2022detecting} take a different approach by introducing self-blended images (SBIs), where pseudo-deepfakes are generated by using the same real image for both the source and the target. This eliminates the need for nearest landmark search and thus makes the pseudo-deepfake generation less computationally expensive. SBI also introduces a set of transformations that produce inconsistencies between the source and target images. Zhao \etal \cite{zhao2021learning} propose an image inconsistency generator (I2G) to synthesize pseudo-deepfakes, in combination with a novel pair-wise self-consistency (PCL) learning approach. Chen \etal \cite{chen2022self_adversarial} propose an adversarial training strategy to dynamically construct pseudo-deepfakes, making them increasingly harder to detect by a given detector. The same authors later propose a one-shot test-time-training (OST) meta-learning approach \cite{chen2022ost}, where pseudo-deepfakes are generated at testing by blending real test and training images and using these to update the current model through one-shot training. Unlike traditional binary classification approaches, pseudo-deepfake-based methods use limited to no fake media during training, which can prevent overfitting to specific manipulations and thus produce more generalizable deepfake detectors.

\noindent\textbf{Anomaly detection based techniques.} Anomaly detection (AD) is a common technique used in machine learning, aimed at identifying patterns or events that significantly deviate from the norm. The goal of AD techniques is to learn representations of only "normal" samples from the training data. Therefore, the assumption is that the AD model is capable of recognizing any normal testing samples as inliers whereas abnormal data is expected to be classified as anomalies. A wide variety of AD techniques are available, including one-class support vector machines (SVMs) \cite{scholkopf1999one_class_svm}, reconstruction-based methods \cite{xia2015reconstruction_based} and GAN-based methods \cite{sabokrou2018adversarially}. AD has achieved great success in areas such as the detection of abnormalities in medical images \cite{baur2019deep_AD_MR} and video surveillance \cite{sultani2018real_video_surv}. A comprehensive list of AD techniques can be found in the survey \cite{yang2022visual}. 

Recently, a small number of publications have adopted AD methods for the deepfake detection task, demonstrating promising generalization performance to unseen manipulations. For example, Khalid \etal \cite{khalid2020ocfakedect} propose OC-FakeDect, a variational autoencoder neural network that is trained to reconstruct only real images. The assumption is that deepfake images should not be reconstructed as effectively as real ones, and thus the reconstruction error can be used as an anomaly score. Larue \etal \cite{larue2023seeable} propose  SeeABLE, which is a method that generates local image perturbations (pseudo-deepfakes) that are then pushed towards predefined prototypes using a regression-based bounded contrastive loss. An anomaly score is then calculated by using the cosine similarity between the trained prototypes and a given test image. Levya \etal \cite{leyva2024data} use a fine-to-coarse Bayesian CNN, trained only on real images, to detect images generated from different GAN and diffusion models. Finally, Meriji \etal \cite{mejri2023untag} introduce UNTAG, which uses a pre-trained backbone to extract deep face embeddings for training an AD model.

\section{Proposed approach}

\begin{figure*}
    \centering
    \includegraphics[width=1.0\linewidth]{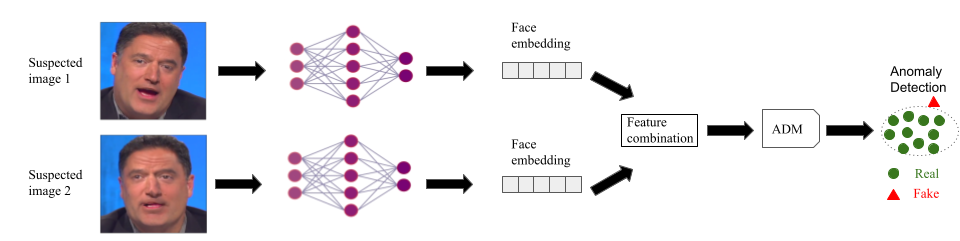}
    \caption{Visualisation of DiffFake. We input pairs of images, corresponding to the same person, to a pre-trained deep neural network to extract the corresponding deep face embeddings. The embeddings are then combined into a single vector, which is then given as input to an ADM trained only on pairs of real images. Finally, the ADM recognizes the initial input as either real or fake.}
    \label{fig:diff_AD}
\end{figure*}

The basis of DiffFake is to detect unnatural changes that may occur between two frames of a video that feature the same person, as graphically depicted in Fig.~\ref{fig:diff_AD}. DiffFake has two main components. The first one is a deep neural network (backbone) that extracts face embeddings from \emph{pairs} of images.  This component is trained using real images and pseudo-deepfakes generated by a novel data augmentation technique. The second component is the anomaly detection model (ADM), which takes as input combined face embeddings, from pairs of images, and outputs whether the input pair is real or fake. The ADM is trained only with features corresponding to real images, extracted by our pre-trained backbone.  In the following sections, we delve into the details of these components.

\subsection{The backbone}

To extract meaningful and generalizable features from facial images, we propose to train the backbone with pseudo-deepfakes as done in previous work \cite{li2020face_x_ray,shiohara2022detecting,sabokrou2018adversarially,zhao2021learning,chen2022ost}. However, unlike these previous methods that introduce either local or global artifacts, we propose to generate pseudo-deepfakes that contain both global and local artifacts through a novel mask generation scheme. This strategy is based on the observation that different types of artifacts can enhance the generalization capability of the backbone, as different deepfake generators usually create videos with different forensic traces \cite{Deepfakes,li2020FaceShifter,NeuralTextures,he2021forgerynet}. 

Specifically, for a given facial image $\bm{I}_t$, we extract its facial landmarks $\bm{L}_t = h(\bm{I_t}) $, where $h: \mathbb{R}^{W \times H \times 3} \rightarrow \mathbb{R}^{68 \times2}$ is a given landmark detector that maps the image $\bm{I}_t$ to a set of $68$ $(x,y)$ coordinates, representing key points on the face. Similar to \cite{shiohara2022detecting}, we use a single real image for both the source and the target. Inconsistencies between the target and source images are introduced by considering a set of \textbf{source-target} transformations $\mathcal{T}_{st}$ that are randomly applied to either image. Given a source and target image, $\bm{I}_t$, $\bm{I}_s$ $\in [0,255]^{W\times H \times 3}$, respectively and a blending mask $\bm{M}$ $\in [0,1]^{W \times
H}$, the blended image is derived as follows:

\begin{equation}
    \bm{I}_B = \bm{I}_s \odot \bm{M} + \bm{I}_t \odot (1-\bm{M}),
    \label{mask}
\end{equation}

\noindent where $\odot$ is the element-wise Hadamard product. 

\begin{figure}
    \centering
    \includegraphics[width=1.0\linewidth]{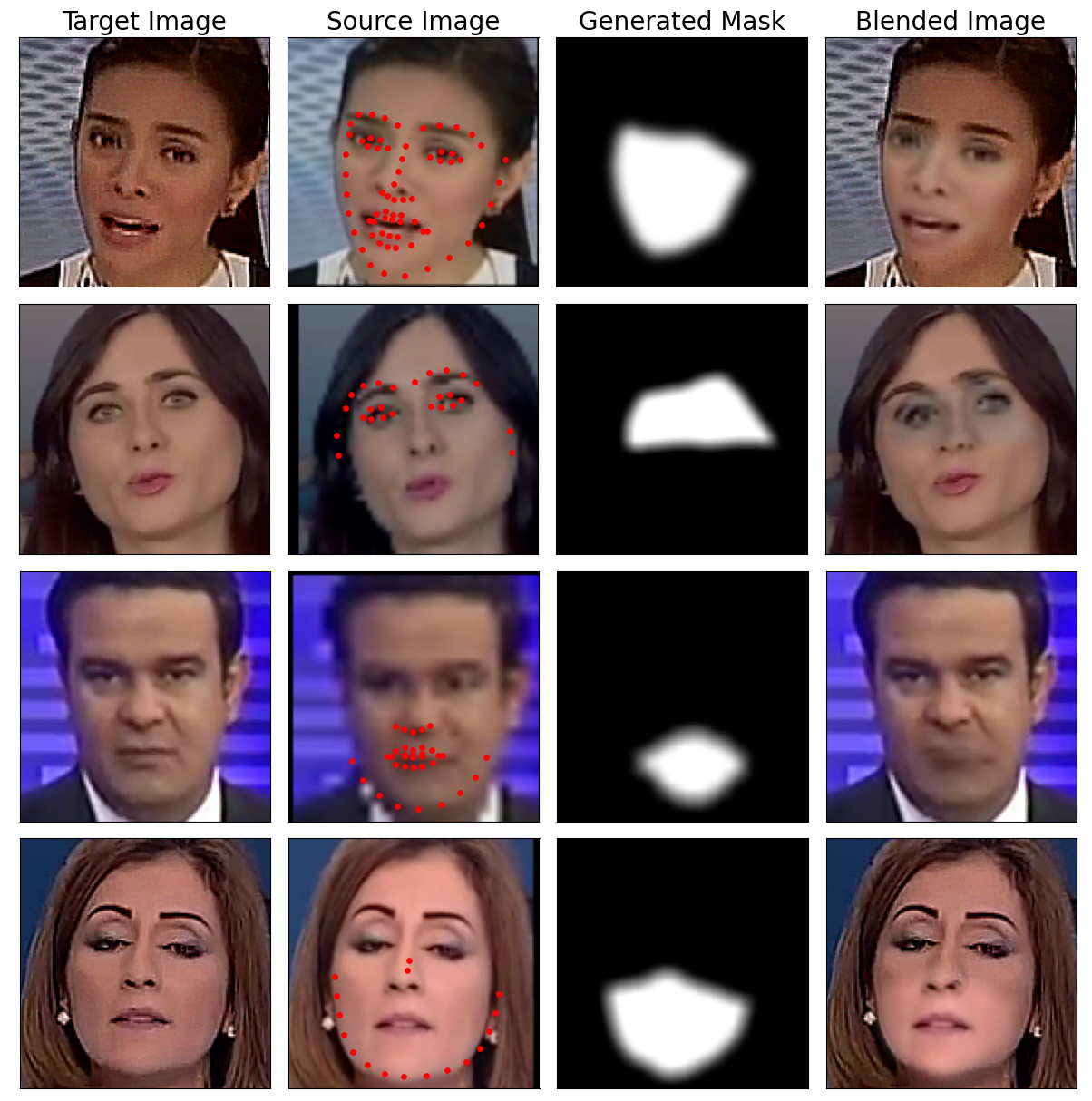}
    \caption{Overview of generating blended images through our proposed method. The second column shows the facial landmarks that are used for the mask generation scheme. The third column shows the resulting masks (after elastic deformation and Gaussian blurring). Finally, the last column showcases the generated pseudo-deepfakes, which contain a variety of artifacts in different parts of the face region.}
    \label{fig:SBI_types}
\end{figure}

Note that in previous work \cite{li2020face_x_ray,shiohara2022detecting,chen2022ost,zhao2021learning}, Eq. \ref{mask} is used to introduce global artifacts in the entire face region, as the blending mask $\bm{M}$ is initialized as the convex hull of the facial landmarks $\bm{L_t}$. In contrast, we expand mask generation to encompass four distinct schemes, each derived from the convex hull of selected subsets of $\bm{L_t}$: (1) all landmarks, (2) the eye region, (3) lower jaw, mouth, and nose apex, (4) the entire jawline and nose tip. As illustrated in Fig. \ref{fig:SBI_types}, these masks allow us to selectively cover the entire face, eyes, mouth, or lower head. To further increase the diversity of the generated pseudo-deepfakes, we modify the shape of the masks by applying elastic deformation and Gaussian smoothing. Furthermore, we vary the blending ratio of the source image, as done in\cite{li2020face_x_ray,shiohara2022detecting,zhao2021learning}. 

Having generated the pseudo-deepfakes, we can train our backbone in a supervised fashion on the task of binary classification. Specifically, given a training set of $N$ images $X = [\bm{x}_0, \bm{x}_1, ..., \bm{x}_{N-1}] $ and their corresponding labels $Y = [y_0, y_1, ..., y_{N-1}]$, we can train a classifier $f: \mathbb{R}^{W \times H \times 3
} \rightarrow \{0, 1\} $ using the binary cross-entropy loss function:

\begin{equation}
    L = -\frac{1}{N} \sum^{N-1}_{i=0} [y_i \log(f(\bm{x}_i) + (1-y_i)\log(1-f(\bm{x}_i))],
\end{equation}

\noindent where $y_i \in \{0,1\}$ indicates the true label of each image $\bm{x}_i$, with $0$ representing a real image and $1$ representing a pseudo-deepfake. After training, we freeze the parameters of the backbone and drop the final classification layer, which allows us to extract deep face embeddings from images.

\subsection{The ADM}

The idea of differential anomaly detection has been used in previous work for identity attack detection \cite{scherhag2020deep_diff_AD,diff_AD}. The goal is to train an ADM only on pairs of real images corresponding to the same person, allowing the model to learn natural changes that can occur between two images of the same subject. Unnatural or extreme changes not observed in real images, such as the ones shown in Fig. \ref{fig:unnatural_changes} (b) and (c), should not be recognized by the model and therefore should be classified as anomalies. It is important to note that differential anomaly detection does not exploit inter-frame dissimilarities between subsequent frames (temporal coherence) as done in \cite{amerini2019optical_flow,sabir2019recurrent}, but rather leverages information extracted from changes of pose or facial expressions. Learning natural changes between frames, where the same person is depicted, requires that the image pairs are sampled from big enough time intervals, such that the expression or head position change to some extent. If we were to choose two subsequent frames from a video to form a pair, the pose of the face would be mostly unchanged.

Given a pre-trained backbone that extracts embeddings of dimension $d$ from an RBG image  $f^{'}: \mathbb{R}^{W \times H \times 3} \rightarrow \mathbb{R}^d$, we can extract face embeddings $\bm{A} = f^{'}(\bm{I}_A)$ and $\bm{B} = f^{'}(\bm{I}_B)$, where $\bm{I}_A$ and $\bm{I}_B$ are images depicting the same person. Similar to \cite{diff_AD}, we propose to fuse the feature vectors $\bm{A}$ and $\bm{B}$ using one of the following feature combinations:

\begin{align}
    &\bm{ABS} = |\bm{A} - \bm{B}| \\
    &\bm{SUB} = \bm{A} - \bm{B} \\
    &(\bm{SUB})^2 = (\bm{A} - \bm{B})^2 \\
    &(\bm{SUB})^3 = (\bm{A} - \bm{B})^3. 
\end{align}

\noindent Then we can use the combined feature vectors, originating only from pairs of real images, to train our ADM.

We propose to use a Gaussian-Mixture-Model (GMM) to model the distribution of combined features as a mixture of $N$ multivariate Gaussian distributions of dimension $d$. Therefore, the probability of an input image $\bm{I}_{t}$ being real ($y=0)$ is given by:

\begin{equation}
    P(y=0|\bm{I_t}) = \sum^{N}_{k=1} \pi_k \mathcal{N}(f^{'}(\bm{I}_t) | \bm{\mu}_k, \bm{\Sigma}_k),
\end{equation}

\noindent where $\pi_k$ represents the mixing coefficient for the $k$-th Gaussian component, with $\sum^{N}_{k=1} \pi_k = 1$, $\mathcal{N}(f^{'}(\bm{I}_t) | \bm{\mu}_k, \bm{\Sigma}_k)$ is the multivariate Gaussian density for the $k$-th component with mean $\bm{\mu}_k$ and covariance $\bm{\Sigma}_k$ and $f^{'}(\bm{I}_t)$ denotes the feature representation of image $\bm{I}_t$ extracted by the backbone.

\section{Experiments}

\subsection{Implementation details}

\noindent\textbf{Image preprocessing.} For a given video, we extract $40$ equally spaced frames, such that frames depict a person's face in different poses and expressions. We use a Haar cascade classifier \cite{haarcascades} and an LBF model \cite{lbfmodel} to extract the bounding boxes and facial landmarks from every frame. If two or more faces are detected in a given image, we choose the one corresponding to the bounding box with the largest area. No further attempt is made to align the faces across frames. Following the protocol of \cite{rossler2019faceforensics++}, we use a conservative crop around each detected face region, by enlarging the bounding boxes by a factor of 1.3. All face-cropped images are normalized using a mean of $[0.5, 0.5, 0.5]$ and standard deviation $[0.5, 0.5, 0.5])$ for each RGB channel. Finally, the images are resized to $224 \times 224$ pixels prior to any of the experiments.

\noindent\textbf{Transformations.} For the source-target transformations, $\mathcal{T}_{st}$, we consider the following augmentations, which are applied randomly: (1) shifting of RGB channels within a range $[-20, 20]$, (2) shifting of HSV channels within a range $[-0.3, 0.3]$, (3) adjusting brightness and contrast by a limit between $[-0.3, 0.3]$, (4) sharpening of image with intensity between $[0.2, 0.5]$, (5) downscaling and then resizing of an image by a factor of either $2$ or $4$. Furthermore, we apply an affine transformation only to the source image to introduce blending boundaries in the resulting blended image. Specifically, the source image is translated along the $x$ and $y$ axes within $\pm 3\%$ of the image and then resized within $\pm 5\%$ of the original size. We select these values to generate a wide variety of subtle artifacts in the blended image, similar to \cite{shiohara2022detecting,larue2023seeable}.

\noindent\textbf{Training of DiffFake.} Training of DiffFake consists of two parts: (1) training the backbone in a supervised fashion using real images and pseudo-deepfakes, (2) training the ADM on combined deep face embeddings extracted from the pre-trained backbone. We choose Efficientnet-b4 \cite{tan2019efficientnet}, pre-trained on ImageNet \cite{deng2009imagenet}, as the backbone of DiffFake. The backbone is trained with the SAM \cite{foret2020SAM} optimizer for $100$ epochs with a batch size of $32$ and an initial learning rate of $0.001$ which is linearly decayed starting from epoch $75$ until the end of training. After training, the last classification layer is discarded, allowing for the extraction of features of dimension $D=1792$. We choose a GMM with $k=3$ clusters (empirically chosen) as the ADM for DiffFake. The GMM is trained on combined feature pairs extracted only from \textbf{real} images, corresponding to the same subject featured in one video. 

\noindent\textbf{Validation} Following the validation protocol of \cite{shiohara2022detecting}, we validate the backbone of our model, during the first phase of training, by constructing a validation set of real and pseudo-deepfake images. This strategy allows us to validate our model without using any fake media. After the binary classification pre-training phase, we choose the backbone that achieves the highest AUC score on the validation set.

\noindent\textbf{Inference.} During the inference phase, we construct 30 randomly selected image pairs for both real and deepfake videos. The anomaly score for individual pairs is computed by calculating the log-likelihood probability under the learned GMM model. Lastly, the anomaly score for the entire video is calculated as the average of the corresponding pair scores.

\subsection{Experimental setup}

\noindent\textbf{Datasets.} In our experiments, we adopt the widely used dataset \textbf{FaceForensics++} (FF++) \cite{rossler2019faceforensics++} for training following the protocol of previous work. The dataset contains 1000 real videos that are split into 720 videos for training, 140 for validation, and 140 for testing. Additionally, FF++ includes 4000 deepfake videos featuring four different manipulation methods: Deepfakes (DF) \cite{Deepfakes}, Face2Face (F2F) \cite{thies2016face2face}, FaceSwap (FS) \cite{FaceSwap} and NeuralTextures \cite{NeuralTextures}. All the videos in FF++ are given in three different qualities corresponding to three distinct compression levels: c0 (no compression), c23 (light compression), c40 (heavy compression).

To assess the performance of DiffFake in cross-dataset scenarios, we adopt four of the most recent deepfake datasets \textbf{Celeb-DF-v2} (CDF) \cite{li2020celeb}, \textbf{Deeperforensics-1.0} (DF1.0) \cite{jiang2020deeperforensics}, \textbf{FaceShifter} (FSh) \cite{li2020FaceShifter},  and \textbf{ForgeryNet} (FNet) \cite{he2021forgerynet}. In all of the datasets we use the recommended splits of the authors for testing, except for FNet where we randomly sampled $480$ videos ($60$ for each manipulation method), as splits are not provided in this particular dataset. Table \ref{table:manipulation_methods} reports the number of real and fake videos used for testing in each dataset and the manipulation method used in each case. It is worth noting that FNet, which is the most recent of the considered deepfake detection datasets, uses 8 different approaches to generate video-level forgeries, making it the most challenging one.

\begin{table}[]
\centering
\begin{tabular}{llll}
\hline
Dataset    & \#Real & \#Fake                   & Manipulation method    \\ \hline
FF++ \cite{rossler2019faceforensics++}      & 140    & \multicolumn{1}{l|}{560} & DF, FS, NT, F2F        \\
CDF \cite{li2020celeb}        & 178    & \multicolumn{1}{l|}{340} & Improved DF            \\
DF1.0 \cite{jiang2020deeperforensics}      & 200    & \multicolumn{1}{l|}{200} & DF-VAE                 \\
FSh \cite{li2020FaceShifter}        & 140    & \multicolumn{1}{l|}{140} & AEI-Net + HEAR-Net     \\
FNet \cite{he2021forgerynet} & 480    & \multicolumn{1}{l|}{480} & 8 different approaches \\ \hline
\end{tabular}
\caption{Details of the deepfake datasets used in our experiments. }%with their associated manipulation methods and the number of real and fake videos they contain in the test set.}
\label{table:manipulation_methods}
\end{table}

\noindent\textbf{Evaluation Metrics.} We report the performance of DiffFake using the Area Under the Receiver Operating Characteristic Curve (AUC), as it is the most commonly used metric in the deepfake detection literature.

\noindent\textbf{SoTA baselines.} We compare the performance of DiffFake to multiple SoTA baselines under various experimental settings. Specifically, (1) two anomaly-detection-based deepfake detection methods, \textbf{UNTAG} \cite{mejri2023untag}, \textbf{OC-FakeDect2} \cite{khalid2020ocfakedect}, (2) four pseudo-deepfake-based methods, \textbf{Face X-ray} \cite{li2020face_x_ray}, \textbf{SLAAD} \cite{chen2022self_adversarial}, \textbf{PCL+I2G} \cite{zhao2021learning}, and \textbf{SBI} \cite{shiohara2022detecting},  (3) the frequency-based methods \textbf{F3Net} \cite{qian2020thinking_in_freq} and \textbf{SRM} \cite{luo2021generalizing_freq}, (4) a binary classification model based on \textbf{Xception} \cite{rossler2019faceforensics++}, and (5) \textbf{RFM}  \cite{wang2021representative}, which is a method that encourages the use of multiple facial regions for forgery detection through forgery attention maps.

To allow for more comprehensive experimental comparisons and to highlight the importance of the differential anomaly detection mechanism in terms of performance, we also introduce a baseline AD method. This baseline extracts deep face embeddings from individual images, using the same pre-trained backbone as DiffFake, and then uses those embeddings to fit an ADM (GMM with $k=3$).

\subsection{Results}

\noindent\textbf{Cross-manipulation evaluation.} An important property of deepfake detectors is their generalization to various manipulation methods. Therefore, we follow the evaluation protocol from \cite{shiohara2022detecting,zhao2021learning} and evaluate the performance of DiffFake on the four manipulation methods from FF++, \ie DF, F2F, FS, and NT. We use the c0 version of the videos for both training and testing to match the experimental setting of the competitors.

Table \ref{table:cross_manipulation_feature_comb}, shows the cross-manipulation evaluation results for DiffFake, for all the different feature combinations. We can observe that there is a very small performance difference across all the considered feature combinations, \ie only $0.4 \%$ difference between the average performances of the highest and lowest entries $(SUB)^3$ and $ABS$. 

Table \ref{table:cross_manipulation} compares the best resulting entry $(SUB)^3$ to the competitors' performance. Our method achieves the best average performance across the four manipulation methods of FF++ with an average AUC score of $99.3\%$. We can see that DiffFake matches or exceeds the performance of the SoTA competitors on all individual datasets. These results show that, in general, our method can effectively detect various types of facial manipulations.

\begin{table}[]
\centering
\begin{tabular}{lllllll}
\hline
\multicolumn{2}{l}{\multirow{2}{*}{Feature Comb.}} & \multicolumn{5}{l}{\;\;\;\;\;\;\;\;\;\;Test Set AUC (\%)} \\ \cline{3-7} 
\multicolumn{2}{l}{}                                     & DF    & F2F   & FS    & NT    & Avg.  \\ \hline
\multicolumn{2}{l}{$ABS$}                                  & 100   & 99.6  & 97.5  & 98.6  & 98.9  \\
\multicolumn{2}{l}{$SUB$}             & 100   & 99.6  & 97.9  & 98.6  & 99.0  \\
\multicolumn{2}{l}{$(SUB)^2$}                                & 100   & 99.6  & 98.2  & 98.6  & 99.1  \\
\multicolumn{2}{l}{$(SUB)^3$}             & 100   & 99.6  & $\bm{98.6}$  & $\bm{98.9}$  & $\bm{99.3}$
\\ \hline
\end{tabular}
\caption{Performance of DiffFake with different feature combinations, under the cross-manipulation setting.}
\label{table:cross_manipulation_feature_comb}
\end{table}

\begin{table}
\centering
\begin{tabular}{lllllll}
\hline
\multicolumn{2}{l}{\multirow{2}{*}{Method}} & \multicolumn{5}{l}{\;\;\;\;\;\;\;\;\;\;\;\;Test Set AUC (\%)} \\ \cline{3-7} 
\multicolumn{2}{l}{}                        & DF    & F2F   & FS    & NT    & Avg.  \\ \hline
\multicolumn{2}{l}{UNTAG \cite{mejri2023untag}}                   & --    & --    & --    & --    & 81.8  \\
\multicolumn{2}{l}{OC-FakeDect2 \cite{khalid2020ocfakedect}}            & 88.4  & 71.2  & 86.1  & 97.5  & 85.8  \\
\multicolumn{2}{l}{Face X-ray \cite{li2020face_x_ray}}              & 99.2  & 98.6  & 98.2  & 98.1  & 98.5  \\
\multicolumn{2}{l}{PCL+I2G \cite{zhao2021learning}}                 & 100   & 99.0  & $\bm{99.9}$  & 97.6  & 99.1  \\
\multicolumn{2}{l}{SBI$^{\dag}$ \cite{shiohara2022detecting}}                     & 99.7  & 99.3  & 98.8  & 98.4  & 99.0  \\ 
\multicolumn{2}{l}{Baseline (ours)}                & 99.6  & 99.3  & 96.8  & 98.2  & 98.5  \\
\multicolumn{2}{l}{DiffFake (ours)}                    & 100   & $\bm{99.6}$  & 98.6  & $\bm{98.9}$  & $\bm{99.3}$ 
\\ \hline
\end{tabular}\caption{Cross-manipulation evaluation results on FF++. DiffFake achieves the best performance on F2F and NT.}\unskip\ignorespaces Note that SBI$^{\dag}$ was evaluated using the official code.
\label{table:cross_manipulation}
\end{table}

\noindent\textbf{Cross-dataset evaluation.} Arguably, the most important quality of deepfake detection algorithms is their generalization to unknown manipulations that can originate from a variety of sources, as this setting mostly resembles real-world situations. Here, we conduct a cross-dataset evaluation where we train our model on FF++ c0 and evaluate on CDF, DF1.0, FNet, and FSh.

In Table \ref{table:cross_dataset_feature_comb}, we present the cross-dataset evaluation results for DiffFake, for all considered feature combinations. In this case, we again see a small but more noticeable average performance difference of $2.3 \%$ between the best and worst performing combinations, \ie, $(SUB)^2$ and $ABS$. Looking at individual datasets like DF1.0, we can see even bigger performance differences of $3.2 \%$ between $(SUB)^2$ and $ABS$, demonstrating that the choice of feature combination can indeed be important for certain datasets. We believe that the performance advantage of $(SUB)^2$ arises from its ability to amplify significant differences while minimizing the impact of minor variations. By squaring the element-wise differences between embeddings $\bm{A}$ and $\bm{B}$, this feature combination emphasizes larger discrepancies, which are often more indicative of a deepfake. At the same time, minor, natural variations—such as those caused by lighting, angle, or pose—are less emphasized when squared, reducing the model's sensitivity to irrelevant noise.

In Table \ref{table:cross_dataset}, we compare DiffFake with $(SUB)^2$ against various competing deepfake detectors. We observe that our method outperforms the SoTA competitors on the DF1.0 and FNet datasets, and nearly matches the best performance in the FSh dataset. In the CDF dataset, DiffFake is outperformed by SBI and SLAAD. However, it is important to note that the latter uses fake videos during training while our method leverages only pseudo-deepfakes.

\begin{table}[]
\centering
\begin{tabular}{lllllll}
\hline
\multicolumn{2}{l}{\multirow{2}{*}{Feature Comb.}} & \multicolumn{5}{l}{\;\;\;\;\;\;\;\;\;\;\;\;Test Set AUC (\%)} \\ \cline{3-7} 
\multicolumn{2}{l}{}                                     & CDF    & DF1.0   & FNet    & FSh    & Avg.  \\ \hline
\multicolumn{2}{l}{$ABS$}                                  & 74.5   & 87.8  & 80.9  & 90.7  & 83.5  \\
\multicolumn{2}{l}{$SUB$}             & 75.1   & 89.8 & 80.0  & 92.4  & 84.3  \\
\multicolumn{2}{l}{$(SUB)^2$}                                & $\bm{76.1}$   & 91.0  & $\bm{83.7}$  & $\bm{92.5}$  & $\bm{85.8}$  \\
\multicolumn{2}{l}{$(SUB)^3$}             & 75.7   & 91.0  & 83.0  & 91.1  & 85.2  
\\ \hline
\end{tabular}
\caption{Performance of DiffFake with different feature combinations, under the cross-dataset setting.}
\label{table:cross_dataset_feature_comb}
\end{table}

\begin{table}[]
\centering
\begin{tabular}{llllll}
\hline
\multicolumn{1}{l|}{\multirow{2}{*}{Method}} & \multicolumn{1}{l|}{\multirow{2}{*}{Real Only}} & \multicolumn{4}{l}{\;\;\;\;\;\;\;\;Test Set AUC(\%)} \\ \cline{3-6} 
\multicolumn{1}{l|}{}                        & \multicolumn{1}{l|}{}                           & CDF        & DF1.0       & FNet      & FSh       \\ \hline
Face X-ray \cite{li2020face_x_ray}                                   & \;\;\;\;Yes                                             & 74.8       & -           & -         & -         \\
SBI$^{\dag}$ \cite{shiohara2022detecting}                                          & \;\;\;\;Yes                                             & $\bm{85.6}$       & 83.3        & 82.2      & 94.0         \\
SLAAD \cite{chen2022self_adversarial}                                        & \;\;\;\;No                                              & 79.7       & 88.9        & -         & -         \\
UNTAG \cite{mejri2023untag}                                        & \;\;\;\;Yes                                             & 74.7       & -           & 77.0      & -         \\
Baseline (ours)                                     & \;\;\;\;Yes                                             & 74.0          & 88.0        & 81.0      & 91.4         \\ 
DiffFake (ours)                                         & \;\;\;\;Yes                                             & 76.1       & $\bm{91.0}$        & $\bm{83.7}$      & 92.5         \\ \hline
\end{tabular}
\caption{Cross-dataset evaluation results on various datasets. DiffFake achieves the best performance on DF1.0 and FNet.}
\label{table:cross_dataset}
\end{table}

\noindent\textbf{Cross-quality evaluation.} In real-world settings, manipulated videos are often post-processed before being posted online. One of the most common post-processing methods is compression, which can eliminate many important artifacts in deepfake videos, thus hindering the performance of deepfake detectors.

Following the experimental protocol of \cite{chen2022self_adversarial}, we evaluate the performance of DiffFake on FF++ with different compression levels. Specifically, we re-train two versions of our model (both the backbone and ADM) with only real videos of FF++ at c23 and c40 compression levels, respectively, and perform testing on DF and FS videos with the same compression levels. Table \ref{table:different_compressions} demonstrates that increasing compression levels can have a significant impact on the performance of deepfake detectors. These results are not surprising since important artifacts introduced by deepfake generators are largely destroyed when the images are highly compressed. Nevertheless, DiffFake retains the highest performance in three out of the four cases showing its robustness under cross-quality settings. We believe that the improvement over the other methods is because DiffFake combines information from pairs of images to infer whether a video is fake or not. This would also explain why DiffFake consistently outperforms the baseline AD method (performance gain ranging from $1.8\%$ to $3.6\%)$, which uses information only from individual frames.

\begin{table}[]
\centering
\begin{tabular}{lllll}
\hline
\multicolumn{1}{l|}{\multirow{3}{*}{Method}} & \multicolumn{4}{l}{\;\;\;\;\;\;Test Set AUC (\%)}                    \\ \cline{2-5} 
\multicolumn{1}{l|}{}                        & \multicolumn{2}{l|}{\;\;\;\;\;\;c40}       & \multicolumn{2}{l}{\;\;\;\;\;\;c23} \\ \cline{2-5} 
\multicolumn{1}{l|}{}                        & DF   & \multicolumn{1}{l|}{FS} & DF         & FS         \\ \hline
Xception \cite{rossler2019faceforensics++}                                     & 58.7 & 51.7                    & 77.0       & 71.8       \\
Face X-ray \cite{li2020face_x_ray}                                   & 57.1 & 51.0                    & 58.5       & 77.9       \\
F3Net \cite{qian2020thinking_in_freq}                                        & 58.3 & 51.9                    & 80.5       & 61.2       \\
RFM \cite{wang2021representative}                                          & 55.8 & 51.6                    & 79.8       & 63.9       \\
SRM \cite{luo2021generalizing_freq}                                         & 55.5 & 52.9                    & 83.8       & $\bm{79.5}$       \\
SLAAD \cite{chen2022self_adversarial}                                        & 62.8 & 56.8                    & 84.6       & 72.1       \\
Baseline (ours)                                     & 74.9 & 55.4                    & 87.9       & 66.1       \\
DiffFake (ours)                                         & $\bm{78.5}$ & $\bm{58.2}$                    & $\bm{89.3}$       & $68.9$       \\ \hline
\end{tabular}
\caption{Cross-quality evaluation results on FS and DF. DiffFake achieves the best performance in three out of the four settings. Note that the results from all other methods are taken from \cite{chen2022self_adversarial}. }
\label{table:different_compressions}
\end{table}

\subsection{Ablation study}

\noindent\textbf{AD vs. classification.} In Table \ref{table:ablation1}, we evaluate the effect of using the differential anomaly detection framework versus standard classification, using DiffFake's backbone as the classifier. We can see that DiffFake has the highest average performance. Specifically, there is a gain of $5.5 \%$, $3.9 \%$, and $4.9 \%$ for DF1.0, FNet, and FSh, respectively. This demonstrates that our proposed strategy of combining a feature extractor with an ADM can lead to significant performance gain across different deepfake datasets.

\begin{table}[t]
\centering
\begin{tabular}{lllll}
\hline
\multirow{2}{*}{Method} & \multicolumn{4}{c}{Test Set AUC (\%)} \\ \cline{2-5} 
                        & DF1.0         & FNet       & FSh        & Avg. \\ \hline
Classification          & 85.5        & 79.8       & 87.6       & 84.3    \\
DiffFake                & $\bm{91.0}$ & $\bm{83.7}$ & $\bm{92.5}$ & $\bm{89.1}$ \\ \hline
\end{tabular}
\caption{Effectiveness of proposed differential anomaly detection framework over standard classification.}
\label{table:ablation1}
\end{table}

\noindent\textbf{Impact of backbone choice.} One of the main components of DiffFake is the backbone, which acts as the feature extractor. In Table \ref{table:ablation2}, we explore the impact of choosing different SoTA network architectures for the backbone, \ie ResNet-50 \cite{he2016resnet}, Xception \cite{chollet2017xception} and EfficientNet-b4 \cite{tan2019efficientnet}, which is our default option. We observe that EfficientNet-b4 outperforms ResNet50 by $5.2 \%$ and Xception by $7.4 \%$, on average . The main difference in average performance is attributed to the DF1.0 and FNet datasets, since all architectures achieve similar performance on FSh. These results demonstrate that larger networks can extract more meaningful features, contributing to the generalization performance of DiffFake.

\begin{table}[]
\centering
\begin{tabular}{lllll}
\hline
\multicolumn{1}{l|}{\multirow{2}{*}{Backbone}} & \multicolumn{4}{l}{\;\;\;\;\;\;\;\;Test Set AUC (\%)}          \\ \cline{2-5} 
\multicolumn{1}{l|}{}                          & DF1.0  & FNet & \multicolumn{1}{l|}{FSh}  & Avg. \\ \hline
ResNet50 \cite{he2016resnet}                                       & 85.8 & 73.4   &  \multicolumn{1}{l|}{92.4} & 83.9      \\
Xception \cite{chollet2017xception}                                       & 80.8 & 75.6 & \multicolumn{1}{l|}{88.8} & 81.7      \\
EfficientNet-b4 \cite{tan2019efficientnet}                                & $\bm{91.0}$ & $\bm{83.7}$ & \multicolumn{1}{l|}{$\bm{92.5}$} & $\bm{89.1}$      \\ \hline
\end{tabular}
\caption{Performance of DiffFake with different backbones.}% The EfficientNet-b4 architecture achieves the best performance across all datasets.}
\label{table:ablation2}
\end{table}

\noindent\textbf{Effect of AD model choice.} In Table \ref{table:ablation3}, we explore the effect of choosing different AD models as the second component of DiffFake, \ie one-class support vector machine (OC-SVM), autoencoder (AE) \cite{sakurada2014anomaly_AE}, and GMM, which is our default option. We observe that, using a GMM as our ADM convincingly outperforms all other models, with an average performance difference of $7.7 \%$  from OC-SVM and $9.1 \%$ from AE. The biggest performance drop in both cases occurs on the DF1.0 dataset, with a performance difference of $18.4 \%$ for OC-SVM and $20.4 \%$ for AE. Surprisingly, both OC-SVM and AE achieve really good performance on the FSh dataset, with the former even surpassing the performance of our default GMM by $0.7 \%$. These results indicate that the probabilistic nature of GMMs is more effective in approximating the distribution of our combined feature data for the deepfake detection task.

\begin{table}[]
\centering
\begin{tabular}{lllll}
\hline
\multicolumn{1}{l|}{\multirow{2}{*}{ADM}} & \multicolumn{4}{l}{\;\;\;\;\;\;\;\;Test Set AUC (\%)}          \\ \cline{2-5} 
\multicolumn{1}{l|}{}                     & DF1.0  & FNet & \multicolumn{1}{l|}{FSh}  & Avg. \\ \hline
OC-SVM                                    & 72.6 & 78.5 & \multicolumn{1}{l|}{$\bm{93.2}$} & 81.4 \\
AE                                        & 70.6 & 78.1 & \multicolumn{1}{l|}{91.4} & 80.0 \\
GMM                                       & $\bm{91.0}$ & $\bm{83.7}$ & \multicolumn{1}{l|}{92.5} & $\bm{89.1}$ \\ \hline
\end{tabular}
\caption{Performance of DiffFake with different ADMs.}%Approximating the distribution of combined features with a GMM yields the best results across all datasets. }
\label{table:ablation3}
\end{table}

\section{Limitations and Future Work}

While the results of DiffFake show promising generalization performance on the cross-manipulation, cross-dataset and degrading video quality settings, our approach does have limitations. For example, DiffFake may be unsuccessful at detecting individual images depicting completely artificial faces, as those generated by SoTA methods like Stable-Diffusion \cite{rombach2022StableDiff} and Midjourney \cite{midjourney}. This limitation is due to the fact that DiffFake works with pairs of images and cannot operate on single frames. Furthermore, very recently text-to-video (T2V) and image-to-video models (I2V), such as Sora \cite{Sora} and Runway-Gen3 \cite{Runway-Gen3}, have become increasingly popular and can create ultra-realistic deepfake videos from a single text prompt or starting frame. DiffFake may also be unsuccessful at detecting deepfakes generated by these models because in our problem we assume that deepfakes are generated by blending two real facial images, whereas T2V and I2V methods generate fully artificial videos. Therefore, a future direction of work is to test and improve the generalization capabilities of DiffFake on the aforementioned cases. Finally, the use of pre-defined feature combinations  to train the GMM model in DiffFake may lead to suboptimal performance. Instead, one can attempt to learn the best feature combination through a multi-layer perceptron designed to find the combination that maximizes the log-likelihood of the underlying GMM model.

\section{Conclusions}

In this paper we proposed DiffFake, a novel deepfake detector that combines differential anomaly detection with pseudo-deepfake generation. The main idea of DiffFake is to leverage pairs of real images corresponding to the same subject, to learn natural changes that can occur between them. Since deepfake videos tend to exhibit unnatural changes between frames, this strategy can effectively detect them. DiffFake uses a feature extractor trained on pseudo-deepfakes generated by a novel data-augmentation technique that introduces both global and local artifacts. Our extensive experiments under various settings showcase that DiffFake can match or even exceed the performance of SoTA competitors.

\noindent\textbf{Acknowledgment.} S. Stamnas was supported by the Engineering and Physical Sciences Research Council through the Mathematics of Systems II Centre for Doctoral Training at the University of Warwick (reference EP/S022244/1).

%%%%%%%%% REFERENCES
{\small
\bibliographystyle{ieee_fullname}
\bibliography{egbib}

\begin{thebibliography}{10}\itemsep=-1pt

\bibitem{Deepfakes}
Deepfakes.
\newblock \url{https://github.com/deepfakes/faceswap}.
\newblock Accessed: 2024-11-01.

\bibitem{FaceSwap}
Faceswap.
\newblock \url{https://github.com/MarekKowalski/FaceSwap/}.
\newblock Accessed: 2024-11-01.

\bibitem{haarcascades}
Haarcascades.
\newblock \url{https://github.com/opencv/opencv/tree/master/data/haarcascades}.
\newblock Accessed: 2024-11-01.

\bibitem{Runway-Gen3}
Introducing gen-3 alpha: A new frontier for video generation.
\newblock \url{https://runwayml.com/research/introducing-gen-3-alpha/}.
\newblock Accessed: 2024-11-01.

\bibitem{lbfmodel}
Lbfmodel.
\newblock \url{https://github.com/kurnianggoro/GSOC2017/blob/master/data/lbfmodel.yaml}.
\newblock Accessed: 2024-11-01.

\bibitem{midjourney}
Midjourney.
\newblock \url{https://www.midjourney.com/}.
\newblock Accessed: 2024-11-01.

\bibitem{Sora}
Video generation models as world simulators.
\newblock \url{https://openai.com/index/video-generation-models-as-world-simulators/}.
\newblock Accessed: 2024-11-01.

\bibitem{afchar2018mesonet}
Darius Afchar, Vincent Nozick, Junichi Yamagishi, and Isao Echizen.
\newblock Mesonet: a compact facial video forgery detection network.
\newblock In {\em 2018 IEEE international workshop on information forensics and security (WIFS)}, pages 1--7. IEEE, 2018.

\bibitem{amerini2019optical_flow}
Irene Amerini, Leonardo Galteri, Roberto Caldelli, and Alberto Del~Bimbo.
\newblock Deepfake video detection through optical flow based cnn.
\newblock In {\em Proceedings of the IEEE/CVF international conference on computer vision workshops}, pages 0--0, 2019.

\bibitem{baur2019deep_AD_MR}
Christoph Baur, Benedikt Wiestler, Shadi Albarqouni, and Nassir Navab.
\newblock Deep autoencoding models for unsupervised anomaly segmentation in brain mr images.
\newblock In {\em Brainlesion: Glioma, Multiple Sclerosis, Stroke and Traumatic Brain Injuries: 4th International Workshop, BrainLes 2018, Held in Conjunction with MICCAI 2018, Granada, Spain, September 16, 2018, Revised Selected Papers, Part I 4}, pages 161--169. Springer, 2019.

\bibitem{bayar2016deep_constrained_layers}
Belhassen Bayar and Matthew~C Stamm.
\newblock A deep learning approach to universal image manipulation detection using a new convolutional layer.
\newblock In {\em Proceedings of the 4th ACM workshop on information hiding and multimedia security}, pages 5--10, 2016.

\bibitem{chen2022self_adversarial}
Liang Chen, Yong Zhang, Yibing Song, Lingqiao Liu, and Jue Wang.
\newblock Self-supervised learning of adversarial example: Towards good generalizations for deepfake detection.
\newblock In {\em Proceedings of the IEEE/CVF conference on computer vision and pattern recognition}, pages 18710--18719, 2022.

\bibitem{chen2022ost}
Liang Chen, Yong Zhang, Yibing Song, Jue Wang, and Lingqiao Liu.
\newblock Ost: Improving generalization of deepfake detection via one-shot test-time training.
\newblock {\em Advances in Neural Information Processing Systems}, 35:24597--24610, 2022.

\bibitem{chen2021local_freq}
Shen Chen, Taiping Yao, Yang Chen, Shouhong Ding, Jilin Li, and Rongrong Ji.
\newblock Local relation learning for face forgery detection.
\newblock In {\em Proceedings of the AAAI conference on artificial intelligence}, volume~35, pages 1081--1088, 2021.

\bibitem{chollet2017xception}
Fran{\c{c}}ois Chollet.
\newblock Xception: Deep learning with depthwise separable convolutions.
\newblock In {\em Proceedings of the IEEE conference on computer vision and pattern recognition}, pages 1251--1258, 2017.

\bibitem{dang2020detection_attention}
Hao Dang, Feng Liu, Joel Stehouwer, Xiaoming Liu, and Anil~K Jain.
\newblock On the detection of digital face manipulation.
\newblock In {\em Proceedings of the IEEE/CVF Conference on Computer Vision and Pattern recognition}, pages 5781--5790, 2020.

\bibitem{deng2009imagenet}
Jia Deng, Wei Dong, Richard Socher, Li-Jia Li, Kai Li, and Li Fei-Fei.
\newblock Imagenet: A large-scale hierarchical image database.
\newblock In {\em 2009 IEEE conference on computer vision and pattern recognition}, pages 248--255. Ieee, 2009.

\bibitem{dickson2021deepfake}
E Dickson.
\newblock Deepfake porn is still a threat, particularly for k-pop stars, 2019.

\bibitem{foret2020SAM}
Pierre Foret, Ariel Kleiner, Hossein Mobahi, and Behnam Neyshabur.
\newblock Sharpness-aware minimization for efficiently improving generalization.
\newblock {\em arXiv preprint arXiv:2010.01412}, 2020.

\bibitem{frank2020leveraging_freq}
Joel Frank, Thorsten Eisenhofer, Lea Sch{\"o}nherr, Asja Fischer, Dorothea Kolossa, and Thorsten Holz.
\newblock Leveraging frequency analysis for deep fake image recognition.
\newblock In {\em International conference on machine learning}, pages 3247--3258. PMLR, 2020.

\bibitem{goodfellow2014GAN}
Ian Goodfellow, Jean Pouget-Abadie, Mehdi Mirza, Bing Xu, David Warde-Farley, Sherjil Ozair, Aaron Courville, and Yoshua Bengio.
\newblock Generative adversarial nets.
\newblock {\em Advances in neural information processing systems}, 27, 2014.

\bibitem{guera2018deepfake_recurrent}
David G{\"u}era and Edward~J Delp.
\newblock Deepfake video detection using recurrent neural networks.
\newblock In {\em 2018 15th IEEE international conference on advanced video and signal based surveillance (AVSS)}, pages 1--6. IEEE, 2018.

\bibitem{haliassos2021lips}
Alexandros Haliassos, Konstantinos Vougioukas, Stavros Petridis, and Maja Pantic.
\newblock Lips don't lie: A generalisable and robust approach to face forgery detection.
\newblock In {\em Proceedings of the IEEE/CVF conference on computer vision and pattern recognition}, pages 5039--5049, 2021.

\bibitem{he2016resnet}
Kaiming He, Xiangyu Zhang, Shaoqing Ren, and Jian Sun.
\newblock Deep residual learning for image recognition.
\newblock In {\em Proceedings of the IEEE conference on computer vision and pattern recognition}, pages 770--778, 2016.

\bibitem{he2021forgerynet}
Yinan He, Bei Gan, Siyu Chen, Yichun Zhou, Guojun Yin, Luchuan Song, Lu Sheng, Jing Shao, and Ziwei Liu.
\newblock Forgerynet: A versatile benchmark for comprehensive forgery analysis.
\newblock In {\em Proceedings of the IEEE/CVF conference on computer vision and pattern recognition}, pages 4360--4369, 2021.

\bibitem{ho2020DDPM}
Jonathan Ho, Ajay Jain, and Pieter Abbeel.
\newblock Denoising diffusion probabilistic models.
\newblock {\em Advances in neural information processing systems}, 33:6840--6851, 2020.

\bibitem{diff_AD}
Mathias Ibsen, L{\'a}zaro~J Gonz{\'a}lez-Soler, Christian Rathgeb, Pawel Drozdowski, Marta Gomez-Barrero, and Christoph Busch.
\newblock Differential anomaly detection for facial images.
\newblock In {\em 2021 IEEE International Workshop on Information Forensics and Security (WIFS)}, pages 1--6. IEEE, 2021.

\bibitem{jiang2020deeperforensics}
Liming Jiang, Ren Li, Wayne Wu, Chen Qian, and Chen~Change Loy.
\newblock Deeperforensics-1.0: A large-scale dataset for real-world face forgery detection.
\newblock In {\em Proceedings of the IEEE/CVF conference on computer vision and pattern recognition}, pages 2889--2898, 2020.

\bibitem{jung2020blinking}
Tackhyun Jung, Sangwon Kim, and Keecheon Kim.
\newblock Deepvision: Deepfakes detection using human eye blinking pattern.
\newblock {\em IEEE Access}, 8:83144--83154, 2020.

\bibitem{karras2019styleGAN}
Tero Karras, Samuli Laine, and Timo Aila.
\newblock A style-based generator architecture for generative adversarial networks.
\newblock In {\em Proceedings of the IEEE/CVF conference on computer vision and pattern recognition}, pages 4401--4410, 2019.

\bibitem{khalid2020ocfakedect}
Hasam Khalid and Simon~S Woo.
\newblock Oc-fakedect: Classifying deepfakes using one-class variational autoencoder.
\newblock In {\em Proceedings of the IEEE/CVF conference on computer vision and pattern recognition workshops}, pages 656--657, 2020.

\bibitem{larue2023seeable}
Nicolas Larue, Ngoc-Son Vu, Vitomir Struc, Peter Peer, and Vassilis Christophides.
\newblock Seeable: Soft discrepancies and bounded contrastive learning for exposing deepfakes.
\newblock In {\em Proceedings of the IEEE/CVF International Conference on Computer Vision}, pages 21011--21021, 2023.

\bibitem{leyva2024data}
Roberto Leyva, Victor Sanchez, Gregory Epiphaniou, and Carsten Maple.
\newblock Data-agnostic face image synthesis detection using bayesian cnns.
\newblock {\em Pattern Recognition Letters}, 183:64--70, 2024.

\bibitem{li2020FaceShifter}
Lingzhi Li, Jianmin Bao, Hao Yang, Dong Chen, and Fang Wen.
\newblock Advancing high fidelity identity swapping for forgery detection.
\newblock In {\em Proceedings of the IEEE/CVF conference on computer vision and pattern recognition}, pages 5074--5083, 2020.

\bibitem{li2020face_x_ray}
Lingzhi Li, Jianmin Bao, Ting Zhang, Hao Yang, Dong Chen, Fang Wen, and Baining Guo.
\newblock Face x-ray for more general face forgery detection.
\newblock In {\em Proceedings of the IEEE/CVF conference on computer vision and pattern recognition}, pages 5001--5010, 2020.

\bibitem{li2020celeb}
Yuezun Li, Xin Yang, Pu Sun, Honggang Qi, and Siwei Lyu.
\newblock Celeb-df: A large-scale challenging dataset for deepfake forensics.
\newblock In {\em Proceedings of the IEEE/CVF conference on computer vision and pattern recognition}, pages 3207--3216, 2020.

\bibitem{liu2021spatial_phase_freq}
Honggu Liu, Xiaodan Li, Wenbo Zhou, Yuefeng Chen, Yuan He, Hui Xue, Weiming Zhang, and Nenghai Yu.
\newblock Spatial-phase shallow learning: rethinking face forgery detection in frequency domain.
\newblock In {\em Proceedings of the IEEE/CVF conference on computer vision and pattern recognition}, pages 772--781, 2021.

\bibitem{luo2021generalizing_freq}
Yuchen Luo, Yong Zhang, Junchi Yan, and Wei Liu.
\newblock Generalizing face forgery detection with high-frequency features.
\newblock In {\em Proceedings of the IEEE/CVF conference on computer vision and pattern recognition}, pages 16317--16326, 2021.

\bibitem{maiano2022depthfake}
Luca Maiano, Lorenzo Papa, Ketbjano Vocaj, and Irene Amerini.
\newblock Depthfake: a depth-based strategy for detecting deepfake videos.
\newblock In {\em International Conference on Pattern Recognition}, pages 17--31. Springer, 2022.

\bibitem{mejri2023untag}
Nesryne Mejri, Enjie Ghorbel, and Djamila Aouada.
\newblock Untag: Learning generic features for unsupervised type-agnostic deepfake detection.
\newblock In {\em ICASSP 2023-2023 IEEE International Conference on Acoustics, Speech and Signal Processing (ICASSP)}, pages 1--5. IEEE, 2023.

\bibitem{oltermann2022european}
Philip Oltermann.
\newblock European politicians duped into deepfake video calls with mayor of kyiv.
\newblock {\em The Guardian}, 2022.

\bibitem{qian2020thinking_in_freq}
Yuyang Qian, Guojun Yin, Lu Sheng, Zixuan Chen, and Jing Shao.
\newblock Thinking in frequency: Face forgery detection by mining frequency-aware clues.
\newblock In {\em European conference on computer vision}, pages 86--103. Springer, 2020.

\bibitem{rombach2022StableDiff}
Robin Rombach, Andreas Blattmann, Dominik Lorenz, Patrick Esser, and Bj{\"o}rn Ommer.
\newblock High-resolution image synthesis with latent diffusion models.
\newblock In {\em Proceedings of the IEEE/CVF conference on computer vision and pattern recognition}, pages 10684--10695, 2022.

\bibitem{rossler2019faceforensics++}
Andreas Rossler, Davide Cozzolino, Luisa Verdoliva, Christian Riess, Justus Thies, and Matthias Nie{\ss}ner.
\newblock Faceforensics++: Learning to detect manipulated facial images.
\newblock In {\em Proceedings of the IEEE/CVF international conference on computer vision}, pages 1--11, 2019.

\bibitem{sabir2019recurrent}
Ekraam Sabir, Jiaxin Cheng, Ayush Jaiswal, Wael AbdAlmageed, Iacopo Masi, and Prem Natarajan.
\newblock Recurrent convolutional strategies for face manipulation detection in videos.
\newblock {\em Interfaces (GUI)}, 3(1):80--87, 2019.

\bibitem{sabokrou2018adversarially}
Mohammad Sabokrou, Mohammad Khalooei, Mahmood Fathy, and Ehsan Adeli.
\newblock Adversarially learned one-class classifier for novelty detection.
\newblock In {\em Proceedings of the IEEE conference on computer vision and pattern recognition}, pages 3379--3388, 2018.

\bibitem{sakurada2014anomaly_AE}
Mayu Sakurada and Takehisa Yairi.
\newblock Anomaly detection using autoencoders with nonlinear dimensionality reduction.
\newblock In {\em Proceedings of the MLSDA 2014 2nd workshop on machine learning for sensory data analysis}, pages 4--11, 2014.

\bibitem{scherhag2020deep_diff_AD}
Ulrich Scherhag, Christian Rathgeb, Johannes Merkle, and Christoph Busch.
\newblock Deep face representations for differential morphing attack detection.
\newblock {\em IEEE transactions on information forensics and security}, 15:3625--3639, 2020.

\bibitem{scholkopf1999one_class_svm}
Bernhard Sch{\"o}lkopf, Robert~C Williamson, Alex Smola, John Shawe-Taylor, and John Platt.
\newblock Support vector method for novelty detection.
\newblock {\em Advances in neural information processing systems}, 12, 1999.

\bibitem{shiohara2022detecting}
Kaede Shiohara and Toshihiko Yamasaki.
\newblock Detecting deepfakes with self-blended images.
\newblock In {\em Proceedings of the IEEE/CVF Conference on Computer Vision and Pattern Recognition}, pages 18720--18729, 2022.

\bibitem{sultani2018real_video_surv}
Waqas Sultani, Chen Chen, and Mubarak Shah.
\newblock Real-world anomaly detection in surveillance videos.
\newblock In {\em Proceedings of the IEEE conference on computer vision and pattern recognition}, pages 6479--6488, 2018.

\bibitem{tan2019efficientnet}
Mingxing Tan and Quoc Le.
\newblock Efficientnet: Rethinking model scaling for convolutional neural networks.
\newblock In {\em International conference on machine learning}, pages 6105--6114. PMLR, 2019.

\bibitem{NeuralTextures}
Justus Thies, Michael Zollh{\"o}fer, and Matthias Nie{\ss}ner.
\newblock Deferred neural rendering: Image synthesis using neural textures.
\newblock {\em Acm Transactions on Graphics (TOG)}, 38(4):1--12, 2019.

\bibitem{thies2016face2face}
Justus Thies, Michael Zollhofer, Marc Stamminger, Christian Theobalt, and Matthias Nie{\ss}ner.
\newblock Face2face: Real-time face capture and reenactment of rgb videos.
\newblock In {\em Proceedings of the IEEE conference on computer vision and pattern recognition}, pages 2387--2395, 2016.

\bibitem{wang2021representative}
Chengrui Wang and Weihong Deng.
\newblock Representative forgery mining for fake face detection.
\newblock In {\em Proceedings of the IEEE/CVF conference on computer vision and pattern recognition}, pages 14923--14932, 2021.

\bibitem{xia2015reconstruction_based}
Yan Xia, Xudong Cao, Fang Wen, Gang Hua, and Jian Sun.
\newblock Learning discriminative reconstructions for unsupervised outlier removal.
\newblock In {\em Proceedings of the IEEE international conference on computer vision}, pages 1511--1519, 2015.

\bibitem{yang2022visual}
Jie Yang, Ruijie Xu, Zhiquan Qi, and Yong Shi.
\newblock Visual anomaly detection for images: A systematic survey.
\newblock {\em Procedia computer science}, 199:471--478, 2022.

\bibitem{yang2019head_poses}
Xin Yang, Yuezun Li, and Siwei Lyu.
\newblock Exposing deep fakes using inconsistent head poses.
\newblock In {\em ICASSP 2019-2019 IEEE International Conference on Acoustics, Speech and Signal Processing (ICASSP)}, pages 8261--8265. IEEE, 2019.

\bibitem{zhang2019detecting_freq}
Xu Zhang, Svebor Karaman, and Shih-Fu Chang.
\newblock Detecting and simulating artifacts in gan fake images.
\newblock In {\em 2019 IEEE international workshop on information forensics and security (WIFS)}, pages 1--6. IEEE, 2019.

\bibitem{zhao2021multi}
Hanqing Zhao, Wenbo Zhou, Dongdong Chen, Tianyi Wei, Weiming Zhang, and Nenghai Yu.
\newblock Multi-attentional deepfake detection.
\newblock In {\em Proceedings of the IEEE/CVF conference on computer vision and pattern recognition}, pages 2185--2194, 2021.

\bibitem{zhao2021learning}
Tianchen Zhao, Xiang Xu, Mingze Xu, Hui Ding, Yuanjun Xiong, and Wei Xia.
\newblock Learning self-consistency for deepfake detection.
\newblock In {\em Proceedings of the IEEE/CVF international conference on computer vision}, pages 15023--15033, 2021.

\end{thebibliography}
}

\end{document}